\documentclass{article}

\usepackage{microtype}
\usepackage{graphicx}
\usepackage{subfigure}
\usepackage{booktabs} 

\usepackage{breakurl}
\usepackage[breaklinks]{hyperref}

\usepackage[accepted]{icml2023}

\usepackage{amsmath}
\usepackage{amssymb}
\usepackage{mathtools}
\usepackage{amsthm}

\usepackage[capitalize,noabbrev]{cleveref}

\theoremstyle{plain}

\theoremstyle{definition}

\theoremstyle{remark}

\usepackage[disable,textsize=tiny]{todonotes}

\icmltitlerunning{Exploring Antitrust and Platform Power in Generative AI}

\begin{document}

\twocolumn[
\icmltitle{Exploring Antitrust and Platform Power in Generative AI}



\begin{icmlauthorlist}
\icmlauthor{Konrad Kollnig}{maas}
\icmlauthor{Qian Li}{maas}
\end{icmlauthorlist}

\icmlaffiliation{maas}{Maastricht University}

\icmlcorrespondingauthor{Konrad Kollnig}{firstname.lastname@maastrichtuniversity.nl}

\icmlkeywords{Generative AI, Competition Law, Antitrust}

\vskip 0.3in
]



\printAffiliationsAndNotice{}  

The concentration of power in a few digital technology companies has become a subject of increasing interest in both academic and non-academic discussions.
One of the most noteworthy contributions to the debate is Lina Khan's \textit{Amazon's Antitrust Paradox} ~\cite{khan_amazons_2017}.
In this work, Khan contends that Amazon has systematically exerted its dominance in online retail to eliminate competitors and subsequently charge above-market prices.
This work contributed to Khan's appointment as the chair of the US Federal Trade Commission (FTC), one of the most influential antitrust organisations.
Today, several ongoing antitrust lawsuits in the US and Europe involve major technology companies like Apple, Google/Alphabet, and Facebook/Meta.
In the realm of generative AI, we are once again witnessing the same companies taking the lead in technological advancements, leaving little room for others to compete.
This article examines the market dominance of these corporations in the technology stack behind generative AI from an antitrust law perspective.

\textbf{Background.} Antitrust law has long been on the books. As early as 1890, the Sherman Antitrust Act sought to restrict anti-competitive and monopolistic corporate behaviour in the US.
It famously led to the break-up of \textit{Standard Oil} in 1911, which was the largest and most influential oil company of its time~\cite{lamoreaux_problem_2019}.
However, US antitrust law had become almost forgotten ever since the Chicago School of economics of the 1970s and the presidency of Ronald Reagan that implemented these thoughts~\cite{wu_curse_2018}.
The Microsoft antitrust case, running from 1998--2001, is seen as the last major antitrust action against a tech company in the US, despite the vastly increased influence of tech companies over society nowadays.
In the European Union, Articles 101--102 TFEU (Treaty on the Functioning of the European Union) provide similar restrictions as in the USA, although the interpretation in practice can differ substantially.

The assessment of market dominance and abuse of this dominance in tech is still an evolving area of scholarship and not finally settled. While there are many pending antitrust cases that are pursued by US and European authorities, no major case has yet been successful and a break-up of any big tech company (except TikTok on the grounds of national security, but not on antitrust), similar to Standard Oil, seems unlikely.
A major challenge remains the long duration that antitrust cases take.
In the European Union, both the \textit{Google Android} (leading to a €4.125bn fine for Google) case and \textit{Google Shopping} case (leading to a €2.42bn fine for Google) took many years to be developed by the European Commission, and then to be settled before the Court of Justice of the European Union (about four years each).
Defining that market and showing the abuse of dominance remains challenging, in light of the use of various sophisticated digital technologies.

\textbf{Analysis: The Generative AI stack.} In generative AI, there are various elements that are part of market success. The most important are access to large amounts of quality data, top talent and expertise, vast financial resources, a suitable infrastructure for development and training, and cutting-edge models trained upon those elements. A further aspect cutting across these is participation in, leadership of, and providing funding for (academic) research.

The foundation of any model is a large source of \textit{data}. To a large extent, generative AI in vision and text has been using data that was sourced from the public domain. Thus far, such data can be gathered with relatively few resources. However, the incumbents still have some competitive edge.
While Microsoft, the owner of GitHub, is unclear about whether it has used private repositories for the training of its Copilot code completion tool, it might~--~in any case~--~have used them for model testing~\cite{butterick_github_2022}.
Similarly, BloombergGPT is an example of using NLP for finance, and training on Bloomberg's proprietary data that has been amassed over many decades~\cite{wu_bloomberggpt_2023}.

Once the relevant data is obtained, a suitable model architecture must be conceived and developed. Since deep learning remains more an engineering discipline rather than actual science with strong theoretical foundations, this requires access to a large number of highly skilled engineers.
The salaries for these engineers are beyond imagination for many individuals, and usually go into six-digit territory, if not beyond.
Only some of the most well-funded companies, like those in tech, are able to pay those salaries.

Significant financial resources are also necessary in order to train the very machine learning models.
According to Sam Altman, the CEO of OpenAI, the cost for training the GPT-3 model ran into the tens of millions of US dollars~\cite{metz_meet_2020}. However, this does not even account for the fact that state-of-the-art LLMs also require much manual human feedback so as to increase the quality of output, e.g. from low-paid workers in Kenya~\cite{billy_perrigo_2_2023}.
Furthermore, the training of such models often happens in proprietary infrastructures (e.g. Amazon AWS or Google Cloud), on proprietary hardware (e.g. Google's TPU), and using industry-dominated frameworks (e.g. Google's TensorFlow).

A further aspect, that's underlying the previous ones, is the fact that a few companies have a major influence on academic research. For example, one study found that 97\% of computer science faculty  with a focus on ethics at top universities had received funding from big tech companies~\cite{abdalla_grey_2021}.
At the top machine learning conference, many reviewers and authors work for those same companies.

In the end, the result is the trained model, of which the most powerful are currently proprietary.
While open-sourcing is currently discussed as a solution, this might also create further risks and may not address the issues of a concentration of capabilities with a few actors in GenAI.

\textbf{Vertical integration in generative AI.} Based upon the above analysis, Google stands out as a company with a high level of vertical integration in the generative AI stack. It might be argued that Google, despite this dominance, has not (yet) managed to deploy an LLM that is able to compete with OpenAI/Microsoft's GPT-4. However, it has also been argued that Google has more to lose in terms of its reputation, being one of the most trusted sources of information on the internet and with generative AI commonly spreading false information.
In American antitrust law, vertical integration has, following the theories of \cite{posner_antitrust_1978} and \cite{bork_antitrust_1978} from the 1970s, not been seen as a concern for intervention since it might be argued that there is no immediate effect on prices (consumer welfare standard). Companies that are vertically, but not horizontally, integrated still have to compete with various other companies in each of the different markets along which this vertical integration takes place.
In a perfect market, this would then, as theorised, not translate into higher prices for consumers.
Apple is the perfect example for such a vertically integrated company, since it controls many market aspects in the iOS ecosystem, stretching from the development of the raw components for iPhones to the development of the iOS operating system.
However, this theory has been questioned, most notably by scholars like \citet{khan_amazons_2017} and \citet{wu_curse_2018}, and is currently being challenged in courts.

\textbf{Competition in generative AI in the future.} As it stands, it is highly difficult for new entrants to develop their own competitive generative AI models.
Historical dominance matters in the race for leadership in generative AI.
It could be argued that large tech companies have been using their dominance in other markets (e.g. online ads or search) to obtain disproportionate profit margins~\cite{competition_and_markets_authority_online_2020}, and use those profits to have a competitive edge in a completely different market, i.e. that around generative AI.
As a result, there is a good chance that we are headed for a future, in which the same incumbents derive vast profits from their historical advantage in tech, and impose a monopoly tax on everyone else.
As knowledge around the development of generative AI becomes more widespread and computing resources continue to become cheaper, the impact of this dominance may lessen somewhat.
However, with current methods, this is unlikely to overcome the current reliance on still significant financial resources, vast amounts of quality data, access to talent, and independent human training and verification.
Indeed, the upcoming EU AI Act may worsen these inequities and reinforce existing market structures. The latest draft as adopted by the EU Parliament includes, among other aspects, stringent obligations for generative AI even if free and open-source, provisions concerning API access to foundation models, extraterritorial scope, and high potential fines for non-compliance~\cite{european_parliament_meps_2023}.
This might then, again, profit those companies that already have vast amounts of resources and may let everyone else pay higher prices than necessary.
It remains to be seen if this law will go ahead as it currently stands (since it just left committee stage), and whether it will have much of a tangible effect or will merely be another paper tiger without much immediate practical relevance like the EU's General Data Protection Regulation (GDPR)~\cite{kollnig_regulatory_2023}.

\textbf{Conclusions and future work.}
The translation between law and tech remains a key challenge in regulating digital ecosystems.
As we continue to produce laws with limited practical understanding of digital technologies, we should not be surprised if this mainly serves the most well-resourced players and goes against the very aims and fundamental rights and freedoms that such laws seek to protect.
In future work, from the perspective of EU competition law,
it would be important to conduct more substantial quantitative analysis into the relevant (product and geographic) markets of specific generative AI systems based on the concept of substitutability, the conditions of access to the defined market, and the extent of market dominance.
Further, the obligations for AI companies under the EU Digital and Markets Services Acts, which include a wealth of obligations for digital platforms, are an interesting field for further study.

\section*{Acknowledgements}

We thank Fazl Barez for helpful discussions.

\bibliography{references}
\bibliographystyle{icml2023}



\end{document}